\documentclass[conference, anonymous=true]{IEEEtran}
\IEEEoverridecommandlockouts
\usepackage{cite}
\usepackage{bbding}
\usepackage{amsmath, amsfonts}
\usepackage{amssymb}
\usepackage{algorithm, algorithmic}
\usepackage{wasysym}
\usepackage{graphicx}
\usepackage{textcomp}
\usepackage{tablefootnote}
\usepackage{latexsym}
\usepackage{xcolor}
\usepackage{booktabs}
\usepackage{multirow}
\def\BibTeX{{\rm B\kern-.05em{\sc i\kern-.025em b}\kern-.08em
    T\kern-.1667em\lower.7ex\hbox{E}\kern-.125emX}}
\begin{document}

\title{Zero-Shot Relational Learning for Multimodal Knowledge Graphs

}

\author{\IEEEauthorblockN{Rui Cai}
\IEEEauthorblockA{
\textit{University of California, Davis}\\
ruicai@ucdavis.edu}
\and
\IEEEauthorblockN{Shichao Pei}
\IEEEauthorblockA{
\textit{University of Massachusetts Boston}\\
shichao.pei@umb.edu}
\and
\IEEEauthorblockN{Xiangliang Zhang}
\IEEEauthorblockA{
\textit{University of Notre Dame}\\
xzhang33@nd.edu}
}

\maketitle

\begin{abstract}
Relational learning is an essential task in the domain of knowledge representation, particularly in knowledge graph completion (KGC). 
While relational learning in traditional single-modal settings has been extensively studied,
exploring it within a multimodal KGC context presents distinct challenges and opportunities. 
One of the major challenges is inference on newly discovered relations without any associated training data. This zero-shot relational learning scenario poses unique requirements for multimodal KGC, i.e., 
utilizing multimodality to facilitate relational learning.
However, existing works fail to support the leverage of multimodal information and leave the problem unexplored. In this paper, we propose a novel end-to-end framework, consisting of three components, i.e., multimodal learner, structure consolidator, and relation embedding generator, to integrate diverse multimodal information and knowledge graph structures to facilitate the zero-shot relational learning.
Evaluation results on three multimodal knowledge graphs demonstrate the superior performance of our proposed method. 
\end{abstract}


\begin{figure*}
    \centering
    \includegraphics[width=0.95\textwidth]{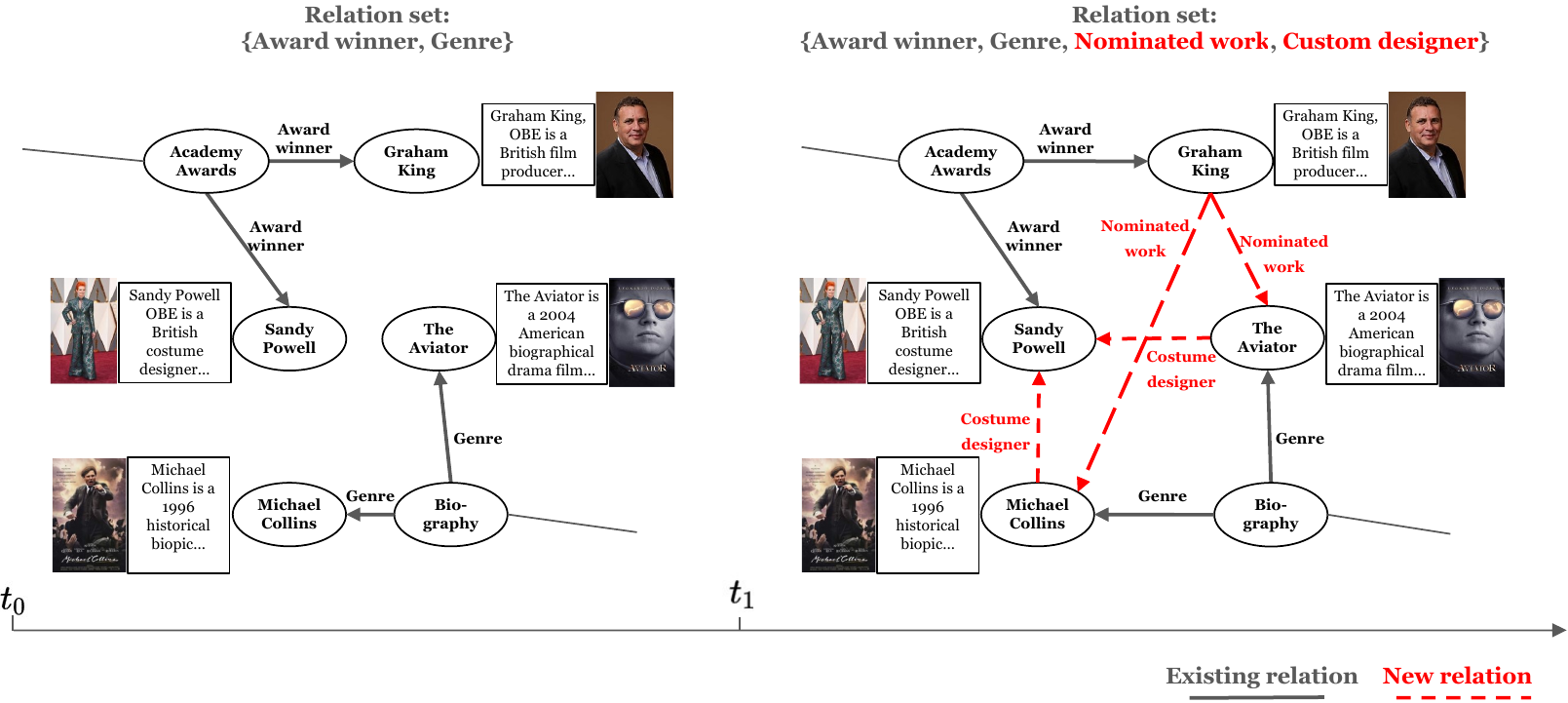}
    \caption{A toy example to illustrate that new relations emerge in the evolution of a multimodal knowledge graph. The MMKG in $t_{0}$ has two branches. After $t_{1}$, two new relations emerge and should be added to the MMKG but without any associated triples.
    %After $t_{1}$ have generated new connections between tail entities after $t_{1}$.
    }
    \label{fig1}
    \vspace{-0.4cm}
\end{figure*}

\section{Introduction}
Knowledge graphs (KGs) have been the prevailing manner for organizing knowledge extracted from different sources and exhibiting relationships between entities as triples in the form of (\textit{head}, \textit{relation}, \textit{tail}), showing widespread applications in natural language processing \cite{pei2019semi, pei2023few} and recommendation systems \cite{surveyrec}. In recent years, multimodal knowledge graphs (MMKGs) \cite{mmkg} have emerged and contain rich multimodal information, such as visuals, text, and structural elements, unfolding more sophisticated capabilities for many tasks, such as named-entity recognition \cite{chen2021multimodal}, and language modeling \cite{endowLMinMMKG}.

Despite the extensive applications of MMKGs, 
the long-tail distribution on relations still severely impedes their use in practice, i.e., a few relations are densely populated with head-tail entity samples but a vast majority have sparse or even no entity association. 
These relations lead to inaccurate learning of the representation due to the lack of sufficient triples. Recent efforts \cite{Gmatching, FAAN, FSRL} %formulate MKGC tasks into the few-shot learning framework and 
attempt to alleviate the impact of long-tail relations in single-modality KGs, %with %the assistance of 
%multimodal information, 
and still require a few triples for each relation and are \textit{incapable of handling the relations without any triples}. 
In reality, the relations without any triples appear frequently in the evolution of MMKGs and these newly discovered relations are to be added to enlarge existing MMKGs, leading to the non-trivial zero-shot scenario, which aims to deduce the relation between entities in an MMKG without any training examples of these relations. 
For instance, in Figure \ref{fig1}, there are two branches in the MMKG at $t_{0}$, one mainly contains award winners of \textit{Academy Awards}, and another mainly contains films of \textit{Biography} genre. After $t_{1}$, two new and uncommon relations emerge during the evolution of the MMKG. As two movies share same costume designer \textit{Sandy Powell}, they are as well the nominated works of \textit{Graham King} in the \textit{Academy Award}. Thus, two new relations \textit{Nominated work} and \textit{Costume designer} are expected to connect entities, resulting in plausible triples. The absence of triples for newly emerged relations, however, makes it difficult for current methods to extrapolate new relations.

Most existing KGC works \cite{transe, transh, transd, distmult, tucker,tang2022positive} only infer the relations with a set of head-tail entity pairs and cannot infer triples for newly discovered relations. 
A few works \cite{zsgan,ontozsl} focus on the zero-shot scenario and require text descriptions extracted from the Web, %as the extra information 
but leave the multi-modal information unexplored. 
In fact, the combination of multimodal information about entities, especially related images, text descriptions, and the original topological properties of KGs, can greatly improve the representation learning of newly discovered relations.
In Figure \ref{fig1}, the description of relation \textit{Custom designer} is that \textit{a professional responsible for designing and creating the overall visual aesthetic and appearance of characters in a film}, while the multimodal information of head entities \textit{The Aviator} and \textit{Michael Collins} contain the shared semantics of film, and the tail entity \textit{Sandy Powell} contains the semantics of designing. As a result, the description of new relations has latent correlations with the multimodal information of entities and the correlation can serve as a guide for inferring missing triples of new relations.
However, modeling the latent correlation is non-trivial due to the diverse modalities of entities and relations.
A straightforward solution is to learn the representation of different modalities separately \textcolor{black}{by using canonical pretrained models like VGG16 \cite{vgg16} and BERT \cite{bert},} and then integrate these embeddings, yet it loses fine-grained semantic information and hard to capture the latent correlation. 
%\textcolor{red}{Furthermore, certain entities within MMKG exhibit incomplete multimodal information, such as absence of images. Existing pretrained models could not sufficiently handle missing modalities.}

To leverage the multimodal information for zero-shot relational learning, we propose a \underline{m}ultimodal \underline{r}elation \underline{e}xtrapolation framework named \textbf{MRE} %(Relational Zero-Shot Masked Autoencoder) 
to learn the representation of newly discovered relations in a zero-shot scenario. 
Specifically, to integrate multimodal information at a fine-grained level, we propose a multimodal learner to encode the multimodal information and model the latent correlation between modalities. 
Then a structure consolidator is employed to incorporate structural information of KGs into the multimodal fusion process and further refine the representation of diverse modalities.
Finally, we design a relation embedding generator to learn accurate relation representations by playing a minimax game following the principle of generative adversarial network \cite{gan}. After training, when faced with a new relation without training triples, the relation representation can be learned using the well-optimized relation embedding generator.
Overall, our contributions in this work include:
\begin{itemize}
    \item We are the first to adopt multimodal information in Multimodal Knowledge Graphs to facilitate relational learning in the zero-shot setting.
    \item We propose a novel end-to-end framework for integrating diverse multimodal information and KG structures to improve relation representation learning. 
    \item We perform extensive experiments on three real-world multimodal knowledge graphs. The experimental results show the superior performance of MRE over the state-of-the-art with significant improvement.
\end{itemize}

\section{Related Work}
% Detailed introduction of related work is given in Appendix \ref{app:related work}.
\subsection{Knowledge Graph Completion}
Traditional knowledge graph completion focuses on utilizing the inherent structural information within knowledge graphs. Its objective is to learn the meaningful representation of entities and relations solely based on the topological characteristics of a knowledge graph. By leveraging the structural information, the goal is to accurately predict missing or potential links between entities in knowledge graphs. TransE \cite{transe}, a typical translational distance-based embedding method, and its variants TransD \cite{transd} and TransH \cite{transh}, aims to minimize the distance between the head entity, relation, and tail entity. Similarily, DistMult \cite{distmult} uses a weighted element-wise dot product to combine two entity embeddings in the embedding space. In recent years, graph neural network (GNN)\cite{gcn} based methods have shown their superior ability to model relational information. RGCN \cite{rgcn} applies GNNs to encode the features of entities by aggregating multi-hop neighborhood information in knowledge graphs. These methods excel in capturing the structural information of knowledge graphs, which inspires us to leverage their insights in the context of multimodal learning.

\textcolor{black}{In order to model multimodal knowledge graphs, IKRL \cite{ikrl} introduces a fusion approach that integrates entity images and structural information using a TransE-like \cite{transe} energy function. Subsequently, TransAE \cite{transae} extends the fusion method to learn both visual and textual knowledge of entities with a multimodal encoder. MKGformer \cite{mkgformer} further proposes a hybrid fusion of multi-level multimodal features. MoSE \cite{mose} considers each multimodal triple as a tightly coupled relation and separates the modalities to learn KG embeddings. Similar to TuckER \cite{tucker}, IMF \cite{imf} employs Tucker Decomposition on each modality to capture interactions between modalities. In contrast to these approaches, our method captures fine-grained multimodal semantics and models the latent correlation to effectively address novel relations in zero-shot settings without requiring relation-specific training triples.}
% Moreover, these methods necessitate a set of training triples for each relation and are incapable of handling new relations in the zero-shot scenario.}

\begin{table}
  \label{comparison}
  \caption{Comparison between previous works and MRE. %Our model combines the fusion strategy in MKGC tasks and lead to better performance on ZSRL.
  (MKGC refers to multimodal knowledge graph completion and ZSRL denotes zero-shot relational learning.)
  }
  \resizebox{0.5\textwidth}{!}
    {
  \begin{tabular}{r|c|c|c|c|c}
      & TransAE & IMF & ZSGAN & OntoZSL & MRE\\
     \hline
    MKGC & \Checkmark & \Checkmark & \XSolid & \XSolid & \Checkmark \\
    \hline
    ZSRL & \XSolid & \XSolid & \Checkmark & \Checkmark & \Checkmark\\ 
    \hline
  \end{tabular}
  }
  \vspace{-0.4cm}
\end{table}

\subsection{Relation Extrapolation}
Few-shot and zero-shot learning have been pervasive paradigms for data-efficient tasks and achieved huge success in many fields \cite{zero-shot, few-shot, zhang2023cross,zhang2022few}. To effectively handle the long-tail relations in KGs, some representative works \cite{Gmatching, FAAN, FSRL, li2024learning} focus on few-shot relational learning in link prediction tasks. Gmatching \cite{Gmatching} studies the one-shot scenario of relations and models entity pairs with their local graph structures. FAAN \cite{FAAN} extends to the few-shot scenario with the consideration of dynamic properties of entities. FSRL \cite{FSRL} uses a relation-aware encoder to encode neighbors of an entity and aggregate multiple entity pairs as supporting information. This methods primarily center around the extrapolation of relations using existing entity pair, but they still rely on the availability of fact triples to accurately infer relations.
Furthermore, few works \cite{zsgan, ontozsl, trgcn} in KGC attempt to predict missing triples for unseen relations in a zero-shot scenario. ZSGAN \cite{zsgan} employs generative adversarial network (GAN) \cite{gan} to extrapolate relations and generates relation embeddings based on the textual description of relations. OntoZSL \cite{ontozsl} incorporates ontology schema as prior knowledge to guide GAN \cite{gan} in inferring unseen relations. TR-GCN \cite{trgcn} models semantic representations for unseen relations based on ontology graph and descriptions. However, \textcolor{black}{in contrast to our model, which utilizes a multimodal encoder to capture the latent semantics between entity pairs and their corresponding relations,} these methods rely on pre-trained language models or well-defined prior knowledge to encode descriptions. Consequently, they fail to fully exploit the inherent latent correlations between entities, relations, and the underlying knowledge graph structure.

\begin{figure*}[h]
  \centering
  \includegraphics[width=\textwidth]{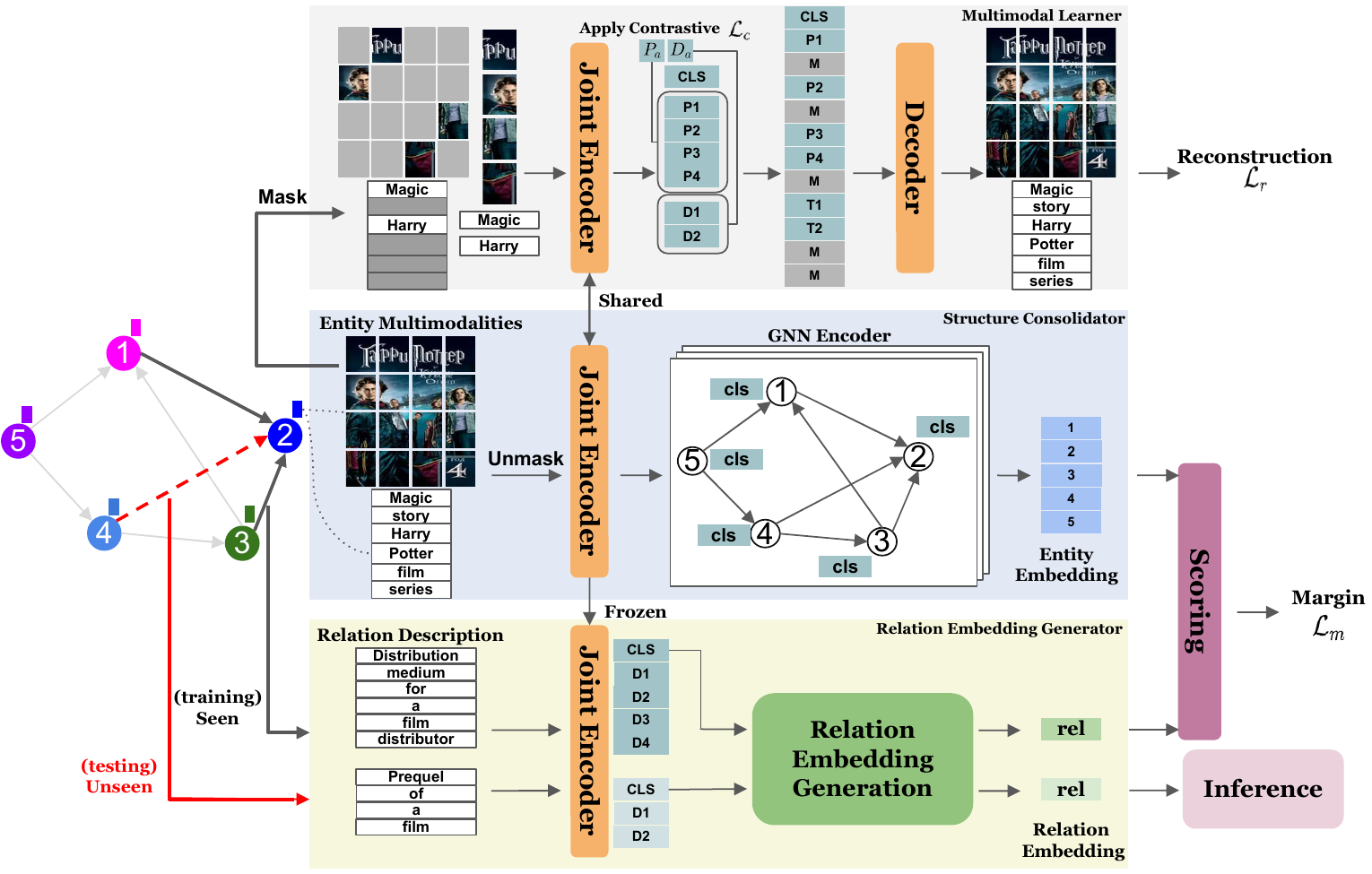}
  \caption{Training pipeline of MRE. \textcolor{black}{The image-and-text pairs of entities are first masked and aligned through a reconstruction procedure at the Multimodal Learner. Then multimodal pairs are unmasked and the \textit{cls} tokens, obtained after Joint Encoder' encoding process, are initialized in the GNN Encoder and fused with KG's topology at the Structure Consolidator. Relation Embedding Generator encodes and generates relation embeddings based on relation descriptions.}}
    \label{trainingpipeline}
    \vspace{-0.5cm}
\end{figure*}

\section{Preliminaries}

\subsection{Multimodal Knowledge Graph Completion}
Given a multimodal knowledge graph $\mathcal{G} = \{ {\mathcal{E}, \mathcal{R}, \mathcal{T},  \mathcal{E}_{m}}\}$, where $\mathcal{E}$ refers to a set of entities, $\mathcal{R}$ denotes a set of relations, and the triple set $\mathcal{T} = \{{(e_h, r, e_t)|e_h \in \mathcal{E}, e_t \in \mathcal{E}, r \in \mathcal{R} }\}$ denotes a set of facts, and $\mathcal{E}_{m} = \{{\mathcal{P}, \mathcal{D}}\}$ includes modalities of entities, i.e., images and text descriptions. The multimodal knowledge graph completion (MKGC) aims to infer the most plausible missing triples from the candidate set 
$\{ ( e_h, r, e_t ) | e_t \in \mathcal{E} \land ( e_h, r, e_t) \notin \mathcal{T} \}$ for each incomplete triple $ ( e_h, r, ? )$ with multimodal information. (or inferring from $\{ ( e_h, r, e_t) | e_h \in \mathcal{E} \land (e_h, r, e_t) \notin \mathcal{T} \}$ for $ (?, r, e_t)$).

% \textcolor{red}{In this task, all entities and relations involve in training and testing.}

\subsection{Multimodal Zero-Shot Relational Learning}
In multimodal relational zero-shot learning, we focus on inferring triples for new relations with the utilization of multimodal information in KGs. Typically, we split the relation set $\mathcal{R}$ into two sets. One is a set of seen relations $\mathcal{R}_{s}=\{r_s|r_s \in \mathcal{R}\}$ and another is a set of unseen relations $\mathcal{R}_{u}=\{r_u|r_u \in \mathcal{R}\}$.
Each relation in $\mathcal{R} = \mathcal{R}_{s} \cup \mathcal{R}_{u}$ has a text description to depict the semantic meaning of the relation.
Then the set of triples $\mathcal{T}_s =\{(e_h, r_s, e_t) | (e_h,  e_t) \in \mathcal{E}, r_s \in \mathcal{R}_{s} \}$ associated with $\mathcal{R}_{s}$ is used for training. And we utilize the set of triples $\mathcal{T}_u =\{(e_h, r_u, e_t) | (e_h,  e_t) \in \mathcal{E}, r_u \in \mathcal{R}_{u} \}$ associated with $\mathcal{R}_{u}$ for testing. Note that in the setting of multimodal KGs, we use the multimodality of each entity in the training and testing.
The target of multimodal zero-shot relational learning is to infer the most plausible missing triples from the candidate set 
$\{ ( e_h, r_u, e_t ) | e_t \in \mathcal{E} \land ( e_h, r_u, e_t) \notin \mathcal{T}_u \}$ for each incomplete triple $ ( e_h, r_u, ? )$, given unseen relation $r_u$.

\section{Methodology}
\label{methodology}
\subsection{Overall Architecture}
The overall framework of MRE, shown in Figure \ref{trainingpipeline}, mainly consists of three modules, Multimodal Learner, Structure Consolidator, and Relational Embedding Generator. Multimodal Learner and Structure Consolidator are unified as a two-stage modality fusion strategy. 
The multimodal learner, including a joint encoder and decoder, fuses visual and textual modalities, then is combined in the Structure Consolidator which is a graph convolutional network \cite{gcn} based module to encode knowledge graph structural information and integrate with other modalities.
%, we use the \textit{cls} token as initial node features, use a GCN-based network and scoring function to fuse the structural information into the fused two modalities. 
Then to generate embedding for relations, the relation embedding generator is designed as a combination of a projector and a discriminator. 
The projector aims to project the encoding of relation descriptions from the joint encoder to relation embeddings.
% %to help fuse structural information in Structure Consolidator based on text descriptions; 
The projector and discriminator play a minimax game 
% In order 
to obtain accurate relation representations.

%the relation discriminator tries to distinguish the fake representation generated from the relation projector with noise and the real representation obtained from the entity pairs following the principle of generative adversarial network.

% generating relations with noise as fake embeddings to involve in generative adverserial networks training, aiming at pushing the relation embeddings closer to the cluster center of its related entity pairs \cite{zsgan}. 
% We adopt a shift training procedure, in which we first train to fuse multimodal knowledge for several epochs and turn to train ZSL part. We want to make the relation embedding generated both suitable in the knowledge graph structural space and latent space of entity-pair after feature extractor's encoding. 

\subsection{Multimodal Learner}
To learn the latent correlation between the modalities of relations and entities, inspired by the mask and reconstruction procedure \cite{mae, m3ae}, we design a multimodal learner that maps visual and textual information into the same feature space and aligns them when two modalities refer to the same element. The feature space depicts the correlation between two modalities and is able to map the relation descriptions to the region close to the related visual and textual elements. Specifically,
%The multimodal learner is to do alignment between visual and textual information and use attention mechanism \cite{transformer} to learn semantic information within the multimodal data that are correlated with structural information in the next stage. 
%we build the multimodal learner based on 
%This module mainly consists of a 
%masked autoencoder \cite{mae}
%and a decoder. To be more specific, we first use neighbor sampling \cite{neighborsampler} method to extract a subgraph of the whole KG, with a group of image-and-text pairs of entities. 
for an entity $e$ with image $P$ and text $D$,  
we first tokenize the input text $D$ into a sequence of discrete tokens $\left\{D_i\right\}^{|D|}_{i=1}$ using BERT tokenizer \cite{bert}. Then each token is converted into a learnable embedding and added with a learnable textual modality embedding and a positional encoding. After that, we obtain the text token embeddings as $\mathbf{D}\in\mathbb{R}^{|D|\times d}$ where $d$ is the embedding dimension of tokens.
%a sequence of text token embeddings as $\left\{\mathbf{d}_i\right\}^{|D|}_{i=1}$. 
%$T_e$ and a positional encoding ${D^p_i}$. 
Then image $P$ is partitioned into regular non-overlapping patches $\left\{P_i\right\}^{|P|}_{i=1}$ of pixels, following Vision Transformers \cite{vit} and projected into a learnable embedding as well, then added with a learnable visual modality embedding and a positional encoding. After that, we can obtain the patch token embeddings as $\mathbf{P}\in\mathbb{R}^{|P|\times d}$. %\left\{\mathbf{p}_i\right\}^{|P|}_{i=1}$. 
Finally, the patch token embeddings $\mathbf{P}$ and the text token embeddings $\mathbf{D}$ are  concatenated into a single embedding matrix $\mathbf{S}_e = \mathbf{P} \oplus \mathbf{D}$, where $\mathbf{S}_e \in \mathbb{R}^{(|P|+|D|) \times d}$.
%$\mathbf{S}_e = \left\{ \mathbf{p}_1, ..., \mathbf{p}_{|P|}, \mathbf{d}_1, ..., \mathbf{d}_{|D|} \right\}$. 

With the modality concatenation, %inspired by the progress of masked autoencoder \cite{mae},
we further propose to fuse two modalities $\mathbf{S}_e$ using a masking strategy \cite{mae}. Specifically, given the concatenated embedding $\mathbf{S}_e$ for entity $e$, we randomly mask $m$ of tokens in each modality and keep $(1-m)$ of token embeddings in $\mathbf{S}_e$ as $\mathbf{S}_e^M$. Then we design a joint encoder $\Phi_E(\cdot)$ to encode $\mathbf{S}_e^M$ and fuse the modalities as:
\begin{equation}
\label{equ:jointencode}
    \mathbf{O}_e^M = \Phi_E(\mathbf{S}_e^M),
\end{equation}
where $\Phi_E(\cdot)$ is a $N$-layer Transformer which adopts multi-head self-attention to fuse visual and textual modalities, and $\mathbf{O}_e^M = \mathbf{CLS}\oplus \mathbf{S}_e^M$
%and $\mathbf{O}_e^M = \mathbf{CLS}\oplus \mathbf{P}^M \oplus \mathbf{D}^M$ 
denotes the updated representation for tokens in $\mathbf{S}_e^M$ and \textit{CLS} token,  %and we reformat the output $\mathbf{O}_e^M$ into a sequential form.
with a little abuse of symbols.

After updating the representation of tokens in $\mathbf{S}_e^M$, we further insert the embeddings of masked tokens $M=\left\{M_1, ..., M_{|M|}\right\}$ back into $\mathbf{O}_e^M$ in the corresponding columns and obtain the recovered $\hat{\mathbf{S}}_e$ which has the same size as $\mathbf{S}_e$,  %and corresponding $\mathbf{S}_e$, %based on their position encoding, 
and then pass the updated embeddings $\hat{\mathbf{S}}_e$ through a decoder $\Phi_D(\cdot)$, which is another $N$-layer Transformer, to reconstruct $\mathbf{S}_e$. 
Specifically, we separate the different modalities in $M$ to masked image patches set $M^P$=$\left\{M_i^P \right\}^{m\times |P|}_{i=1}$ and masked language tokens set $M^D$=$\left\{M^D_i\right\}^{m\times |D|}_{i=1}$ based on their original positions in $\mathbf{S}_e$. 
Then the decoder works by:

\begin{equation}
%\hat{P}_i,\hat{M}_{I_i},\hat{D}_i,\hat{M}_{D_i}  = \Phi_D(\hat{\mathbf{S}}_e) 
\hat{P}, \hat{D} = \Phi_D(\hat{\mathbf{S}}_e).
\end{equation}

We only consider the masked tokens $\hat{M}^P$ and $\hat{M}^D$ in $\hat{P}$ and  $\hat{D}$ for reconstruction, and we define loss functions for $\hat{M}^{P}$ and $\hat{M}^{D}$ as:
\begin{equation}
    \begin{aligned}
    & \mathcal{L}_{r}^{P} = \frac{1}{|M^P|}\sum\limits_{i=1}^{|M^P|}\Vert \hat{M}_{i}^P-M_{i}^P \Vert ^2 \text{,}\\
    & \mathcal{L}_{r}^{D} = \frac{1}{|M^D|}\sum\limits_{k=1}^{|M^D|}\sum\limits_{c}y_{kc}\log(\frac{\hat{M}^D_{kc}}{\sum_{j}\hat{M}^D_{kj}}) \text{,}
    \end{aligned}
\end{equation}
%$\mathcal{S}_e=\left\{I_1, \hat{MI}_1, ..., I_{|I|}, D_1, \hat{MD}_1, ..., D_{|D|}\right\}$
where %$N_I$, $N_D$ represents the number of $M_I$ and $M_D$, and 
%$\hat{M}_{I_{i_j}}$, ${M_{I_{i_j}}}$ denote the predicted and original $j_{th}$ position of $M_{I_i}$, and ${\hat{M}_{D_{k_j}}}$ is the $j_{th}$ predicted logits of $M_{D_k}$ and 
$y_{kc}$ is an indicator that equals 1 if the prediction is the correct token else 0. \textcolor{black}{$c$ is the class of token with respect to the vocabulary of BERT \cite{bert} tokenizer.} We use hyperparameters $\lambda_p$ and $\lambda_d$ as weights of loss for each modality, then define the whole reconstruction loss as follows:
\begin{equation}
   \mathcal{L}_{r} = \lambda_p * \mathcal{L}_{r}^{P} + \lambda_d * \mathcal{L}_{r}^{D}.
\end{equation}

In addition, to better extract the common features within the multimodal pair, we apply contrastive learning \cite{constrastive,yu2018walkranker} to the two modalities. Specifically, we use image-and-text pairs of $N$ entities as input and each obtains $\mathbf{O}_e^M$ after the joint encoder with encoded visual tokens set $\mathbf{P}_e$ and textual tokens set $\mathbf{D}_e$. We calculate the mean embedding in the two modality sets of each entity $i$ and get the averaged two tokens $\mathbf{P}^{i}_{a}$ and $\mathbf{D}^{i}_{a}$ of $i$ to represent its two modalities. Then we apply a contrastive loss on $N$ entities as:

\begin{equation}
    \mathcal{L}_{c} = -\frac{1}{N}\sum\limits_{i=1}^{N}\log\left[\frac{\exp(s_{i,i} / \tau)}{\Sigma_{j} \exp(s_{i,j} / \tau)}\right]\text{,}
\end{equation}
where $s_{i,j}=|\mathbf{P}^{i}_{a}|^T|\mathbf{D}^{j}_{a}|$ represents the similarity between visual token of entity $i$ and textual token of entity $j$ and $\tau$ is the temperature parameter.

\subsection{Structure Consolidator}
Since the multimodal learner only learns to integrate the visual and textual modalities and the structural modality plays a significant role in the knowledge graph tasks and has not been incorporated, we propose to consolidate the structural modality of KGs into the multimodal learner. 
Specifically, for an entity $e$, we use the token embedding $\mathbf{S}_e$ without mask operations as the input of the joint encoder $\Phi_E(\cdot)$ and use the output embedding $\mathbf{CLS}$ of \textit{CLS} token from the output $\mathbf{O}_e$ as the feature of entity $e$. Similar to previous GNN works \cite{gcn, rgcn}, we then utilize a two-layer aggregation network to encode entity features with neighboring information as:  
\begin{equation}
    \mathbf{x}_e^{(l+1)} = \sigma(\sum\limits_{r\in\mathcal{R}_s}\sum\limits_{j\in\mathcal{N}_{e}^{r}}\frac{1}{|\mathcal{N}_{e}^{r}|}
    W_r^{(l)}\mathbf{x}_j^l+ W_0^{(l)}\mathbf{x}_e^l)\text{,}
\end{equation}
where $\mathbf{x}_e^l$ denotes the representation of entity $e$ at layer $l$ and we use $\mathbf{CLS}$ to initialize $\mathbf{x}_e^0$,  $\mathcal{N}_{e}^{r}$ represents the neighbors of $e$ connected by $r$. $W_r$ and $W_0$ are learnable parameter matrices. \textcolor{black}{$\sigma$ is an activation function for which we use LeakyReLU.}

After the update of entity representations, for a triple $(e_h, r, e_t)$, we can obtain embedding $\mathbf{x}_h$ for $e_h$ and $\mathbf{x}_t$ for $e_t$. 
For relation $r$, we employ the joint encoder $\Phi_E$ to encode its description $D_r$ and pass the relation representation $\mathbf{x}_r$ through a projector (a fully connected network) to match the dimension of $\mathbf{x}_h$, $\mathbf{x}_t$. Then we compute the plausibility score based on TransE \cite{transe} using $f(e_h, r, e_t) = \Vert \mathbf{x}_h + \mathbf{x}_r - \mathbf{x}_t \Vert$ in the scoring module and define a loss $\mathcal{L}_{m}$ as follows:
\begin{equation}
    \mathcal{L}_{m} = \sum\limits_{\substack{(e_h, r, e_t)\in R_s \\ (e_h', r, e_t')\notin R_s}} \left[ \gamma+f(e_h, r, e_t)-f(e_h', r, e_t')\right]_{+}\text{,}
\end{equation}
where $f(e_h', r, e_t')$ are negative triples generated by randomly replacing head or tail entities. $\gamma$ is a pre-defined margin parameter. % to maintain a control distance between the true scores and the false scores.
Note that in the consolidator, the joint encoder will be updated jointly using loss $\mathcal{L}_{m}$, $\mathcal{L}_{r}$ and $\mathcal{L}_{c}$.

\subsection{Relation Embedding Generator}
Relation description $D_r$ of relation $r$ are initially encoded in $\Phi_E$ with frozen gradients and without masking. We first use the BERT tokenizer to tokenize $D_r$, and use the \textit{CLS} token after encoding to directly pass through a projector $\Phi_P(\cdot)$ to get relation embeddings and join in the scoring module as before mentioned. After training the fusion modules for several epochs, we turn to train our zero-shot module with a feature extractor $\Phi_{FE}(\cdot)$, the projector $\Phi_P(\cdot)$, and an extra discriminator $Dis(\cdot)$.

\subsubsection{Feature Extractor}
In the training stage, each relation is associated with a set of triples. We expect the triples of the same relation to be aggregated and separated from different relations. Thus, we treat the entity pairs <\textit{head}, \textit{tail}> in associated triples as space coordinates in high-dimensional latent space and utilize a feature extractor $\Phi_{FE}(\cdot)$ to map entity pairs into a single embedding space. The notation $\Psi(\cdot)$ includes a fully connected (FC) network. $\Phi_{FE}(\cdot)$ is constructed with three fully connected network $\Psi_1$, $\Psi_2$ and $\Psi_3$.
Specifically, we denote $\mathcal{T}_r$ 
%$\mathcal{T}_r = \left\{ (e_h, \hat{r}, e_t)| \hat{r}=r, (e_h, e_t) \in \mathcal{E} \right\}$ 
as a set of triples associated with relation $r$. To represent relation $r$, we first propose an entity encoder $\Psi_1$ to encode the entity embeddings \textcolor{black}{$\mathbf{x}_h$, $\mathbf{x}_t$} in a triple %learned from former modules 
and the representation of $r$ can be obtained as follows:

\begin{equation}
 \mathbf{x}_r = \sigma(\Psi_1(\textcolor{black}{\mathbf{x}_h})\oplus \Psi_1(\textcolor{black}{\mathbf{x}_t}))\text{,}
\end{equation}
where $\sigma$ is the tanh activation function, $\oplus$ is the concatenation operator. Then we use another neighborhood encoder $\Psi_2$ to encode one-hop neighbors of entities in a triple. For entity $e_i$, its original embedding $\mathbf{x}_i$ is updated by:

\begin{equation}
  \textcolor{black}{\mathbf{x}_{i}^{new}} = \frac{1}{|\mathcal{N}_{i}|}\sum_{e_{n} \in \mathcal{N}_{i}} \Psi_2(\mathbf{x}_{n})\text{,}
\end{equation}
where $\mathcal{N}_i$ refers to one-hop neighbors of entity $e_n$ \textcolor{black}{and $\mathbf{x}_n$ is the embedding of $e_n$ obtained from the previous Structure Consolidator.} 
After the neighbor encoding, we obtain the embedding $\mathbf{x}_h^{new}$ and $\mathbf{x}_t^{new}$ for head entity $e_h$ and tail entity $e_t$, then we concatenate the embeddings together and adopt %another FC layer $f_{c3}$ 
$\Psi_3$ to align the dimension of concatenated embedding with \textcolor{black}{embeddings derived from Structure Consolidator} as:
\begin{equation}
\label{fml:phi3}
 \mathbf{x}_{(h, t)} = \Psi_3(\textcolor{black}{\mathbf{x}_{h}^{new}} \oplus \mathbf{x}_{r} \oplus \textcolor{black}{\mathbf{x}_{t}^{new}})\text{,}
\end{equation}

We randomly choose a subset of triples $\{e_h^*,\hat{r},e_t^*\}$ from $\mathcal{T}_r$ as the reference set $\mathcal{O}_r$, mark the rest triples in $\mathcal{T}_r$ as positive set $\mathcal{O}_p=\{e_h^+,\hat{r},e_t^+\}$ and randomly replace the tail entity to obtain a negative set $\mathcal{O}_n=\{e_h^+,\hat{r},e_t^-\}$. We calculate the embedding of each entity pair in three sets using the feature extractor and apply cosine similarity $f(\cdot)$ to compute the score between the target embedding and the embedding from a selected triple $t_p \in \mathcal{O}_p$ or $t_n \in \mathcal{O}_n$. The target embedding is $\textcolor{black}{\mathbf{x}_{tar}} = \frac{1}{|\mathcal{O}_r|}\Sigma \mathbf{x}_{(h^*,t^*)}$ and we define $\mathcal{L}_{f}$ as:

\begin{equation}
    \mathcal{L}_{f} = \frac{1}{|\mathcal{O}_p|}\sum\limits_{\substack{t_p\in \mathcal{O}_p \\ t_n \in \mathcal{O}_n}} \left[ \gamma_{f} + f(\textcolor{black}{\mathbf{x}_{tar}}, \Phi_{FE}(t_p)) - f(\mathbf{x}_{tar}, \Phi_{FE}(t_n))\right]_{+}\text{,}
\end{equation}
where $\gamma_{f}$ is a pre-defined margin parameter. %to control the distance between positive and negative scores.

\subsubsection{Generative Adversarial Training}
GAN \cite{gan, wgan} usually contains a generator to generate plausible embeddings with noise as extra information, and a discriminator $Dis(\cdot)$ to distinguish true embedding from the generated one. In our model, the generator consists of a noisy layer $\Phi_{N}(\cdot)$ and a projector $\Phi_P(\cdot)$. After encoding description $D_r$ using the joint encoder, we input the embedding $\mathbf{CLS}$ of \textit{CLS} token  into $\Phi_{N}$ with an additional noise $\mathbf{z}$ sampled from Gaussian distribution $\mathcal{N}(0, 1)$ to obtain a noisy embedding $\mathbf{x}_{r_n}$ as: 
\begin{equation}
    \mathbf{x}_{r_n} = \Phi_{N}(\mathbf{CLS} \oplus \mathbf{z}).
\end{equation}

Then we use $\mathbf{x}_{r_n}$ as the input of $\Phi_P$, and obtain the fake embedding $\textcolor{black}{\mathbf{x}_{fa}} = \Phi_P(\mathbf{x}_{r_n})$ that will involve in GAN training. For relation $r$, the true embedding is calculated by $\textcolor{black}{\mathbf{x}_{tr}}=\frac{1}{|\mathcal{O}|}\Sigma \mathbf{x}_{(h,t)}$ with all its related entity pairs, since $\textcolor{black}{\mathbf{x}_{tr}}$ serves as the cluster center in feature space. The loss of generator is defined as:
\begin{equation}
    \mathcal{L_G} = -\mathbb{E}\left[Dis(\textcolor{black}{\mathbf{x}_{fa}})\right] + \mathcal{L}_{cls}(\textcolor{black}{\mathbf{x}_{fa}}) + \mathcal{L}_p\text{,}
\end{equation}
where $\mathcal{L_G}$ consists of a Wasserstein loss \cite{wgan} to erase the mode collapse problem, a cross-entropy classification loss to attenuate the inter-class discrimination by classifying $\textcolor{black}{\mathbf{x}_{fa}}$, and a pivot regularization loss \cite{pivot} to push $\textcolor{black}{\mathbf{x}_{fa}}$ closer to the center of cluster. The discriminator takes $\textcolor{black}{\mathbf{x}_{tr}}$ and $\textcolor{black}{\mathbf{x}_{fa}}$ as input, then applies a loss as:

\begin{equation}
    \mathcal{L}_{D} = \mathbb{E}\left[Dis(\mathbf{x}_{fa})\right] - \mathbb{E}\left[Dis(\mathbf{x}_{tr})\right] + \lambda\mathcal{L}_{cls}(\mathbf{x}_{tr})\text{,}
\end{equation}
where the first two terms are the Wasserstein distance \cite{wgan} between the distribution of $\mathbf{x}_{tr}$ and $\mathbf{x}_{fa}$, and $\mathcal{L}_{cls}$ is a binary classification loss to distinguish $\mathbf{x}_{tr}$ from $\mathbf{x}_{fa}$. $\lambda$ is a hyperparameter correlated with $\mathcal{L}_{cls}$. 

\begin{algorithm}[t]
\renewcommand{\algorithmicrequire}{\textbf{Input:}}
\renewcommand{\algorithmicensure}{\textbf{Output:}}
\caption{Training Pipeline of MRE}
\label{alg:pipeline}
\begin{algorithmic}[1]
    \REQUIRE Multimodal Knowledge Graph $\mathcal{G}$, Split train and test set $R_s$, $R_u$ of seen and unseen relations, Images and Text descriptions file $P$, $D$, a preset training epoch $p$
    \ENSURE Trained Model MRE
    \WHILE {not converge}
    \FOR{$i = 1, 2, \cdots, p$}
        \STATE Sample a batch of triples from $R_s$ based on Neighbor Sampler to build a subgraph $\mathcal{G}_{s}$
        \STATE Input $P$, $D$ of entities in $\mathcal{G}_{s}$ into $\Phi_E$, $\Phi_D$ in Multimodal Learner, compute the loss $\mathcal{L}_c, \mathcal{L}_r$
        \STATE Use $\Phi_E$, $\Phi_G$ encoded entity embeddings and $\Phi_E$, $\Phi_P$ encoded relation embeddings to compute the loss $\mathcal{L}_m$
        \STATE Train $\Phi_E$, $\Phi_D$, $\Phi_G$ and $\Phi_P$ by minimizing $\mathcal{L}_1$
    \ENDFOR
    \STATE Train the feature extractor $\Phi_{FE}$ by minimizing $\mathcal{L}_{f}$
    \STATE Train $\Phi_N$, $\Phi_P$ and $Dis$ by maximizing $\mathcal{L}_D$ and minimizing $\mathcal{L}_G$
    \ENDWHILE
    \RETURN MRE
\end{algorithmic}
\end{algorithm}

\subsection{Optimization and Inference}
We first train the fusion module in preset $e_p$ epochs with a combined loss function:
\begin{equation}
    \mathcal{L}_{1} = \lambda_c * \mathcal{L}_{c} + \lambda_r * \mathcal{L}_{r} + \lambda_m * \mathcal{L}_{m},
\end{equation}
then we turn to train zero-shot module in Relation Embedding Generator with losses $\mathcal{L}_{f}$, $\mathcal{L}_{G}$ and $\mathcal{L}_{D}$. The specific procedure is shown at Algorithm \ref{alg:pipeline}.

Once the module is well trained, it serves as a function to map the description embedding from multimodal KG semantic space to the latent space of entity-pair embeddings. For a new triple $(e_h, r_u, e_t)\in\mathcal{T}_u$ with its related multimodal information, and a newly-added relation $r_u$, we first use Multimodal Learner, Structure Consolidator and feature extractor to encode the entity-pair $\mathbf{x}_{(h, t)}$, then we use $\Phi_E$, $\Phi_N$ and $\Phi_P$ to encode the relation embedding and obtain $\mathbf{x}_{fa}$ based on its description $D_r$ and a random noise $\mathbf{z}$. Meanwhile, we use $N$ noise for relation embedding generation and calculate the average to eliminate the bias, as $\mathbf{x}_{fa}=\frac{1}{N}\sum_{i=1}^{N}\mathbf{x}_{fa}^{i}$. 

For the purpose of evaluating the zero-shot ability of MRE, we construct a candidate list $C_{(e_h, r_u)}$ for each triple in the testing set, which are all fake entity pairs with replaced tail entities, and we rank the true tail entity among fake entity pairs in $C_{(e_h, r_u)}$. We calculate the cosine similarity between $\mathbf{x}_{(h, t)}$ and $\mathbf{x}_{fa}$ as the score.

\begin{table}
 \centering
  \caption{Statistics of datasets.}
  \label{tab:datasets}
  \resizebox{0.45\textwidth}{!}{
  \begin{tabular}{ccccc}
    \toprule
    Dataset & Entities & Triples & Relations & Train$\backslash$Val$\backslash$Test\\
    \midrule
    FB15K-237-ZS & 14208 & 280447 & 235 & 245548$\backslash$17303$\backslash$17596\\
    DB15K-ZS & 12741 & 97251 & 157 & 89530$\backslash$2068$\backslash$5653 \\
    WN18-IMG-ZS & 40204 & 149086 & 18 & 117468$\backslash$15900$\backslash$15718 \\
    \bottomrule
  \end{tabular}}
\end{table}
\vspace{-0.2cm}

\begin{table*}
\caption{Zero-shot relational learning results on three datasets. The best results are in bold, and the strongest baseline is indicated with *. \textcolor{black}{\textit{BoW} uses pretrained word embeddings with bag-of-words to encode relation descriptions and \textit{Bert} uses BERT \cite{bert} encoded relation descriptions. \textcolor{black}{Results are averaged across 5 random seeds.}}}
  \label{tab:results}
  \resizebox{\textwidth}{!}{
  \begin{tabular}{ccccccccccccc}
    \toprule
     & \multicolumn{4}{c}{FB15K-237-ZS} & \multicolumn{4}{c}{DB15K-ZS} & \multicolumn{4}{c}{WN18-IMG-ZS}\\
    Model & MRR & Hits@1 & Hits@5 & Hits@10 & MRR & Hits@1 & Hits@5 & Hits@10 & MRR & Hits@1 & Hits@5 & Hits@10 \\
    \midrule
    ZS-TransE(Bert) & 0.013 & 0.004 & 0.016 & 0.023 & 0.017 & 0.010 & 0.015 & 0.023 & 0.007 & 0.001 & 0.004 & 0.009 \\
    ZS-DistMult(Bert) & 0.066 & 0.031 & 0.096 & 0.121 & 0.074 & 0.047 & 0.093 & 0.110  & 0.012 & 0.002 &  0.011 & 0.020\\
   \midrule
    TransE+ZSGAN(BoW) & 0.111 & 0.053 & 0.151 & 0.234 & 0.264 & 0.154 & 0.372 & 0.489 & 0.214 & 0.123 & 0.294 & 0.410 \\
    Distmult+ZSGAN(BoW) &  0.141 & 0.071 & 0.207 & 0.301 & 0.284 & 0.194 & 0.403 & 0.554 & 0.225 & 0.138 & 0.352 & 0.474 \\
    \midrule
    TransAE+ZSGAN(BoW) & 0.153 & 0.085 & 0.224 & 0.310 & 0.307 & 0.209 & 0.458 & 0.563 & 0.359 & 0.233 & 0.401 & 0.498 \\
    % MKGFormer+ZSGAN(BoW)  &  &  &  &  &  &  &  &  \\
    IMF+ZSGAN(BoW)  & 0.177 & 0.094 & 
    $0.249^*$ & 0.351 & 0.340 & 0.213 & 0.439 & $0.592^*$ & $0.388^*$ & 0.274 & 0.456 & 0.530 \\
    \midrule
    TransAE+ZSGAN(Bert) & 0.156 & 0.088 & 0.228 & 0.317 & 0.312 & 0.210 & 0.446 & 0.570  & 0.343 & 0.230 & 0.412 & 0.505 \\
    % MKGFormer+ZSGAN(Bert)  &  &  &  &  &  &  &  &  \\
    IMF+ZSGAN(Bert)  & $0.192^*$ & $0.103^*$ & 0.243 & $0.368^*$ & $0.344^*$ & $0.215^*$ & $0.463^*$ & 0.582 & 0.385 & $0.297^*$ & $0.473^*$ & $0.544^*$\\
    \midrule
    MRE & \textbf{0.211} & \textbf{0.128} & \textbf{0.282} & \textbf{0.379} & \textbf{0.355} & \textbf{0.221} & \textbf{0.511} & \textbf{0.618} & \textbf{0.396} & \textbf{0.316} & \textbf{0.475} & \textbf{0.556}\\
    \midrule
    Improv. ($\%$) & 9.8$\%$ & 28.0$\%$ & 13.2$\%$ & 3.0$\%$ & 3.1$\%$ & 3.7$\%$ & 10.3$\%$ & 4.3$\%$ & 2.8$\%$ & 6.4$\%$ & 0.4$\%$ & 2.2$\%$ \\
    \bottomrule
  \end{tabular}
  }
  \label{table:results}
  \vspace{-0.3cm}
\end{table*} 

\section{Experiments}

\subsection{Datasets}
We adopt three multimodal knowledge graphs FB15K-237, DB15K and WN18-IMG to construct datasets \textbf{FB15K-237-ZS}, \textbf{DB15K-ZS} and \textbf{WN18-IMG-ZS} for evaluating our model in the zero-shot scenario. FB15K-237 is a subset of Freebase \cite{freebase} with 20 images for each entity and a paragraph of description for all entities. DB15K is a subgraph extracted from DBpedia \cite{dbpedia} and WN18-IMG is an extended dataset of WN18 \cite{wn18}, where each provides with 10 images for each entity respectively.
We also randomly sample one image of each entity as the visual modality to make image-text pairs.
Moreover, we split the dataset into two sets in terms of relations. 
One represents a set of seen relations $R_s$, marked as $\mathcal{T}_s$ and another denotes a set of unseen relations $R_u$, marked as $\mathcal{T}_u$. The statistics of the three datasets is shown in Table \ref{tab:datasets}. All entities in unseen set $\mathcal{T}_u$ are included in seen set $\mathcal{T}_s$.

\subsection{Baselines and Metrics}
To demonstrate the zero-shot learning ability of our model, we compare our model with three groups of methods. Note that there is no existing work on zero-shot relational learning on multimodal KGs and we design simple strategies to enable these methods to extrapolate relations. (1) \textbf{ZS-TransE} and \textbf{ZS-DistMult}: We empower TransE \cite{transe} and DistMult \cite{distmult} with the zero-shot ability by using BERT \cite{bert} to encode relation descriptions as the relation representation.
(2) \textbf{TransE+ZSGAN} and \textbf{DistMult+ZSGAN}: We combine ZSGAN \cite{zsgan} with TransE and DistMult to enable traditional KG embedding approaches the capability to extrapolate relations.
(3) \textbf{TransAE+ZSGAN} and \textbf{IMF+ZSGAN}: We integrate ZSGAN with two multimodal KG embedding models TransAE \cite{transae} and IMF \cite{imf} to enable multimodal KG embedding approaches the capability to extrapolate relations. \textcolor{black}{Here, TransAE \cite{transae} trains the knowledge graph embeddings similar to TransE \cite{transe}, but with the addition of visual and textual features of entities. These embeddings are then utilized in conjunction with the ZSGAN framework. Lastly, we consider the IMF+ZSGAN model as a potential state-of-the-art baseline. The IMF model \cite{imf} currently achieves SOTA performance in multimodal knowledge graph link prediction tasks. It combines three modalities (visual, textual, and KG structure) using tucker decomposition and a separated modality scoring method. In our evaluation, we use the output of the last Mutan (Multimodal Tucker Fusion) layer in the IMF model as the trained embeddings. In TransE, Distmult, and TransAE, the dimension of pre-trained embeddings are 200, while in IMF we choose 256 according to their released codes. We do not use ontology-based methods as baselines because of the lack of corresponding ontology for MMKGs.}
It is worth noting that \textbf{TransE+ZSGAN},  \textbf{DistMult+ZSGAN}, \textbf{TransAE+ZSGAN} and \textbf{IMF+ZSGAN} use the bag-of-words method based on pre-trained word embeddings following ZSGAN \cite{zsgan} to generate representations of relation descriptions, while our model MRE directly utilizes the well-trained joint encoder as the text encoder. \textcolor{black}{To ensure fairness, we also use BERT \cite{bert} encoded embeddings of relation descriptions with an embedding length of 768 in baselines as comparison.}
In addition, we adopt two metrics: Mean Reciprocal Ranking (\textit{MRR}) and \textit{Hit@$k$}, and the $k$ we set to 1, 5, 10.

\subsection{Implementation Details}
For each dataset, we partition the seen relations $R_s$ into train and validation relations (i.e. train/val/test:196/10/29 for FB15K-237-ZS, 128/10/19 for DB15K-ZS, 11/3/4 for WN18-IMG-ZS.) 
% The experiments are conducted on a server equipped with an Intel Xeon Gold 6226R CPU and four NVIDIA GeForce RTX 3090ti GPUs, utilizing PyTorch 1.13.1. 
The experiments are conducted on a server equipped with four NVIDIA GeForce RTX 3090ti GPUs, utilizing PyTorch 1.13.1. 
The model parameters are initialized using Xavier initialization and optimized using the Adam \cite{adam}
% \cite{adam}
optimizer with cosine annealing warm restarts scheduler \cite{cosinewarm}.
% \cite{cosinewarm}.
We employ a masked autoencoder with 12 layers of blocks \cite{transformer} in the encoder and 8 layers of blocks in the decoder. The embedding dimension of the tokens in the masked autoencoder is set to 384. During data processing, text descriptions of entities are tokenized into sequences of length 64, and each entity image is divided into patches of size 16x16, resulting in a total of 256 tokens per image. Relation descriptions are tokenized into sequences of length 320.

\begin{table*}
\centering
  \caption{Evaluation results of ablation study on three datasets. The bold numbers indicates the best results among all settings.}
   \label{tab:ablation}
  \resizebox{0.95\textwidth}{!}{
  \begin{tabular}{ccccccccccccc}
    \toprule
    & \multicolumn{4}{c}{FB15K-237-ZS} & \multicolumn{4}{c}{DB15K-ZS} & \multicolumn{4}{c}{WN18-IMG-ZS}\\
     Settings & MRR & Hits@1 & Hits@5 & Hits@10 & MRR & Hits@1 & Hits@5 & Hits@10 & MRR & Hits@1 & Hits@5 & Hits@10\\
    \midrule
    MRE (S + V + T) & \textbf{0.211} & \textbf{0.128} & \textbf{0.282} & \textbf{0.379} & \textbf{0.355} & \textbf{0.221} & \textbf{0.511} & \textbf{0.618} & \textbf{0.396} & \textbf{0.316} & \textbf{0.475} & \textbf{0.556}\\
    Structural + Visual (S + V) & 0.158 & 0.088 & 0.230 & 0.306 & 0.296 & 0.176 & 0.423 & 0.541 & 0.348 & 0.242 & 0.421 & 0.503\\
    Structural + Textual (S + T) & 0.179 & 0.090 & 0.257 & 0.334 & 0.322 & 0.210 & 0.449 & 0.554 & 0.385 & 0.290 & 0.464 & 0.537\\
    w/o REG Module & 0.096 & 0.061 & 0.103 & 0.143 & 0.102 & 0.079 & 0.122 & 0.139 & 0.157 & 0.079 & 0.203 & 0.332\\
    w/o GNN module & 0.199 & 0.107 & 0.261 & 0.348 & 0.331 & 0.208 & 0.465 & 0.587 & 0.377 & 0.295 & 0.434 & 0.541\\
    w/o $\mathcal{L}_{c}$ & 0.206 & 0.114 & 0.268 & 0.343 & 0.312 & 0.188 & 0.444 & 0.572 & 0.390 & 0.311 & 0.468 & 0.544 \\
    \bottomrule
  \end{tabular}
  }
  \vspace{-0.3cm}
\end{table*}

\subsection{Performance Comparison}
\begin{figure}[t]
 \centering
  \includegraphics[width=0.45\textwidth]{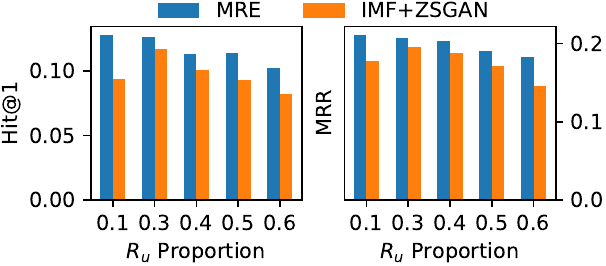}
  \caption{Different $R_s$ and $R_u$ spilt ratios and result comparison of MRE and IMF+ZSGAN regarding $\textit{MRR}$ and $\textit{Hit@1}$ in FB15K-237-ZS. %The left blue bars refer to $MRR$ and $Hit@1$ matrics of MRE and the right orange bars are related metrics of IMF$\&$ZSGAN.
  }
  \label{fig:dif-split}
  \vspace{-0.3cm}
\end{figure}

The overall performance of all metrics on three datasets shown in Table \ref{table:results} demonstrates that our model outperforms all the baselines \textcolor{black}{with both the bag-of-words method and BERT encoder.} Results also show that the performance of \textbf{MRE} is improved significantly compared with the monomodal method \textbf{TransE+ZSGAN} because the multimodality provides rich information and enhances the inference ability to extrapolate new relations. Compared with \textbf{IMF+ZSGAN}, our model achieves different improvement on all metrics, resorting to the joint encoder's effective alignment with description of relations and multimodal information in KGs. 
Also, comparing monomodal models with multimodal models, we can see that adding multimodal information about entities can greatly boost the performance of KGC.  
We also randomly split the dataset and evaluate our model and \textbf{IMF+ZSGAN} on the different splits of $R_s$ and $R_u$ by increasing the proportion of $R_u$ to the ratio 0.3, 0.4, 0.5, 0.6. 
%(We increase the proportion of $R_u$ into ratio 0.3, 0.4, 0.5, 0.6) 
The comparison results are shown in Figure \ref{fig:dif-split}. The results indicate that our model has superior zero-shot learning ability in different splits.

\begin{figure}[t]
    \centering
    \includegraphics[width=0.45\textwidth]{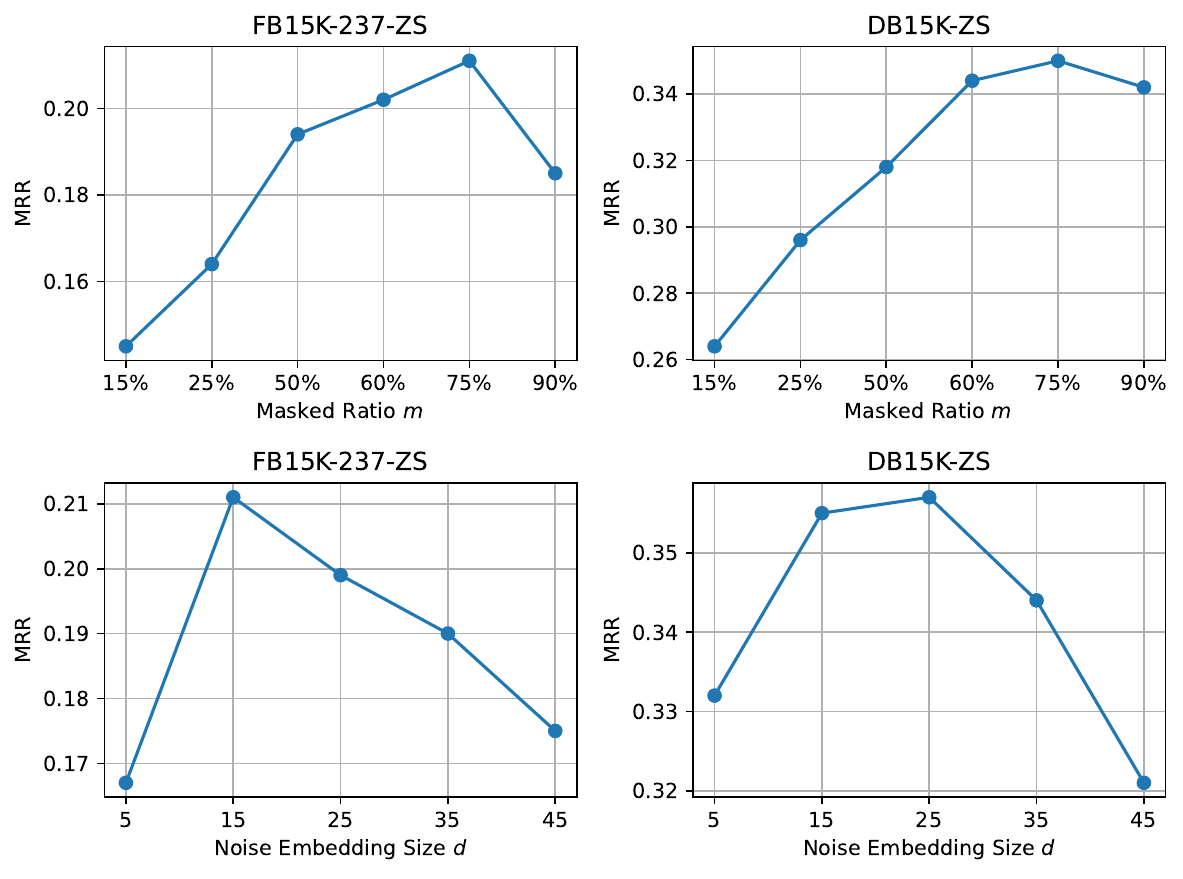}
    \caption{Comparing MRE with different masked ratio $m$ and different noise embedding size $d$. All results are derived from the trained models which achieve best results on validation datasets.}
    \label{fig:parameteranalysis}
    \vspace{-0.4cm}
\end{figure}

\subsection{Ablation Study}
\label{ablationtext}
To gain deeper insight into the effectiveness of each component in the proposed model, we conduct ablation studies by comparing the following variants with \textbf{MRE}: \textcolor{black}{
\textbf{structure + image (S+V)}, \textbf{structure + text (S+T)}, \textbf{structure + image + text (S+V+T)}, \textbf{train without GNN module (w/o GNN module)}, \textbf{train without relation embedding generation module (w/o REG module)}. The Multimodal Learner and Structural Consolidator are combined components designed for multimodal fusion. By performing ablation studies on different modality combinations, we aim to assess their individual and collective contributions to the overall performance of the model. } Note that we do not evaluate the variant simply using structure modality since the joint encoder in our model requires extra modalities to learn the representation of relations. %our MAE encoder need to learn extra modality's information to encode the relation description. 

The results on three graph datasets are summarized in Table \ref{tab:ablation}, %we train MRE with the above 3 different combinations and display the result. 
we can see that the performance of our model \textbf{MRE (S+V+T)} is notably improved when using multimodal information. 
%The model \textbf{MRE (S+V+T)} leverages three modalities and achieves the best zero-shot performance. 
The results prove that modeling the latent correlation between the modalities facilitates the zero-shot relation extrapolation.
%visual information serves as a supplement to enhance the textual semantic information
%, which reaches an agreement with former works \cite{mkgformer,mose}.
\textbf{MRE} outperforms \textbf{S+V}, 
%Textual modality contributes more than visual modality, 
probably due to the reason that our joint encoder is trained solely on images, and the visual modality is not aligned with the textual modality. Therefore,  \textbf{S+V} is unable to accurately encode relation descriptions. We can observe that the performance of \textbf{S+T} is greatly improved compared to \textbf{S+V} because the misalignment between visuals and text is alleviated. 

We also evaluate the effectiveness of REG module, contrastive loss $\mathcal{L}_{c}$ in Multimodal Learner, %to test its contribution in aligning the visual and textual modalities, 
and the GNN encoder of Structure Consolidator. %to assess its capability in learning topological information of KG. 
\textcolor{black}{The results show that the minimax game-based zero-shot training framework inside REG module significantly enhances MRE's zero-shot learning capabilities and $\mathcal{L}_{c}$ has benefited the modality fusion of our model.} Instead of using graph convolutional layers, we use linear layers instead. Compared to the fully connected layers, the designed method has made a huge improvement in all evaluation metrics, indicating the effectiveness of GNN \cite{rgcn} in capturing the structural modality of multimodal knowledge graphs. 

\begin{table*}[t]
\caption{The \textit{MRR} and \textit{Hit@1} of different models with best 5 performance of unseen relations achieved in MRE. Here number represents the related triples of certain relations in the test dataset $R_u$.}
\label{tab:casestudy}
\resizebox{\textwidth}{!}{
\begin{tabular}{cccccccccc}
\toprule

\multirow{2}{*}{Relation} & \multirow{2}{*}{Number} & \multicolumn{2}{c}{\textcolor{black}{MRE(w/o REG)}} & \multicolumn{2}{c}{TransE+ZSGAN} & \multicolumn{2}{c}{IMF+ZSGAN} & \multicolumn{2}{c}{MRE} \\
\cmidrule(r){3-4}\cmidrule(r){5-6}\cmidrule(l){7-8}\cmidrule(l){9-10}
& & MRR & Hit@1 & MRR & Hit@1 & MRR & Hit@1 & MRR & Hit@1 \\
\midrule
$football\_historical\_roster\_position(A)$ & 102 & 0.778 & 0.706 & 0 .637 & 0.451 & 0.995 & 0.990 & 1.000 & 1.000\\
$military\_combatant\_group(B)$ & 711 & 0.575 & 0.528 & 0.741 & 0.658 & 0.652 & 0.541 & 0.846 & 0.782\\
$track\_contribution\_role(C)$ & 2035 & 0.072 & 0.049 & 0.374 & 0.280 & 0.238 & 0.106 & 0.513 & 0.336 \\
$featured\_film\_locations(D)$ & 1170 & 0.106 & 0.042 & 0.315 & 0.223 & 0.189 & 0.035 & 0.482 & 0.381 \\
$recording\_contribution\_role(E)$ & 499 & 0.447 & 0.278 & 0.272 & 0.178 & 0.323 & 0.114 & 0.473 & 0.228 \\
\bottomrule
\end{tabular}
}
\end{table*}

\begin{figure*}
    \centering
    \includegraphics[width=0.75\textwidth]{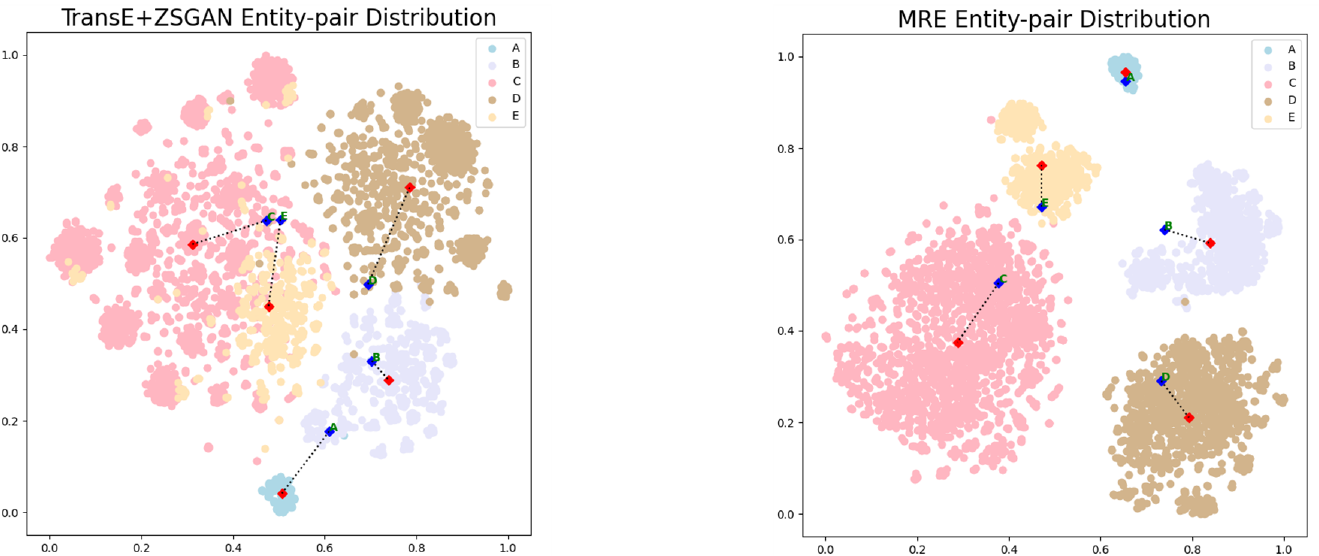}
    \caption{Two figures demonstrate the embedding distribution of selected 5 relations and their related entity pairs in two models after t-SNE analysis. Blue points are generated relation embeddings and red points are cluster centers of each cluster.}
    \label{cluster}
    \vspace{-0.3cm}
\end{figure*}

\subsection{Parameter Analysis}
We investigate the performance influence of the masked ratio $m$ and noise embedding size $d$. The results are presented in figure \ref{fig:parameteranalysis}. We performed evaluation on two datasets while keeping the noise embedding size fixed at 15. We varied the masked ratio from 15$\%$ to 90$\%$. The results suggest that MRE achieves better performance when the masked ratio is relatively high, around 75$\%$. We also evaluate MRE's performance by varying $d$ from 5 to 45 on two datasets while maintaining a fixed masked ratio of 0.75. The results indicate that larger embedding sizes can lead to overfitting issues and the optimal embedding size falls within the range of 15 to 25.

\subsection{ Time Complexity Analysis and Comparison} 
The Multimodal Learner in MRE mainly focuses on fusing information from image-text pair of each entity. Suppose each image has pixel size $H * W$ and is divided by the encoder into pixels with size $p * p$, the number of patches is $N_I = \frac{H * W}{p^2}$. The embedding size of Multimodal Learner is $d$, and the number of transformer layers in encoder is $N_E$, we consider the complexity of image encoding as $O(N_E*(N_I^2 * d + N_I * d^2))$ according to \cite{transformer}, denoted as $O(I)$. Similarly, the length of text tokens is $N_D$, and the complexity of text encoding is $O(N_E*(N_D^2 * d + N_D * d^2))$, denoted as $O(D)$. Primarily, the encoder of multimodal learner has a higher complexity compared to the decoder, so we omit the decoder here. The second part is Structural Consolidator where we fused three modalities.  
We apply $N_{gcn}$ GCN layers to fuse structural modality, which has a complexity of $O(N_{gcn}*(N_{rel} + N_{ent})*d^2)$ according to \cite{rgcn}. $N_{rel}$ and $N_{ent}$ are numbers of relations and entities in a batch. 
The Relational Embedding Generator has a complexity of $O(N_{rel} * D)$, and the overall approximate time complexity for each step of MRE is $O(N_{ent}*(I+D) + N_{rel} * D)$ as multimodal fusion part are most complex. According to IMF, its complexity is $O(N_{ent} * (I+D))$ while it doesn't consider encoding relation descriptions. Therefore, our model and SOTA model IMF have the same level of time complexity.

\subsection{Case Study}
\label{app:casestudy}
To evaluate the quality of generated relation embeddings, we assess MRE's performance on the unseen dataset $R_u$ of FB15K-237-ZS. Specifically, we select the top 5 relations based on their \textit{MRR} scores, comparing other baselines including \textbf{MRE}, \textbf{IMF+ZSGAN}, \textbf{TransE+ZSGAN}, and \textcolor{black}{\textbf{MRE(w/o REG)}}.
% , which is the model setting discussed in \ref{ablationtext}.} In this evaluation, we focus on fusing different modalities of entities and training the projector using the margin ranking loss $\mathcal{L}_m$.
Table \ref{tab:casestudy} presents the results for the specific relations.

The numbers indicate the count of related triples for each relation in $R_u$.
Results demonstrate MRE's superior performance in generating relation embeddings compared to baselines. 
Notably, baselines tend to exhibit good zero-shot learning performance on relations with fewer triples (i.e. A, B, E), which leverage the learned experience from the seen relation dataset but struggle with predicting unseen relations that involve more triples (i.e. C, D). In contrast, MRE demonstrates consistent performance across all relations regardless of their associated triple count.   

To gain insights into the learned representations of extrapolated relations in MMKGs and their proximity to the cluster center of related entity pairs, we utilize t-SNE for visualizing the relation embeddings and the embeddings of their associated entity pairs in a high-dimensional space. This analysis focuses on the 5 selected relations in Table \ref{tab:casestudy}. Figure \ref{cluster} presents two visualizations: one for the baseline model \textbf{TransE+ZSGAN} and the other for MRE. In terms of embedding distributions, MRE exhibits greater intra-class similarity and smaller inter-class similarity compared to the baseline. This is evident from the distinct clustering of the 5 relations in MRE, contrasting with the more scattered clusters in the baseline. Importantly, the relation embeddings generated by MRE demonstrate closer proximity to their respective cluster centers, contributing to its superior zero-shot performance compared to the baseline. 
Furthermore, the results highlight the advantage of incorporating multimodal information into zero-shot relational learning. In monomodal model TransE+ZSGAN, the cluster distributions appear chaotic. In contrast, MRE leverages multimodal information to form more compact and cohesive clusters in the visualized space.

\vspace{-0.4cm}
\section{Conclusion}

In this paper, we proposed a novel model called MRE (Multimodal Relation Extrapolation) for inferring missing triples of newly discovered relations for MMKGs in the zero-shot scenario.
Specifically, we designed a multimodal learner to map visual and textual modalities into the same feature space and model the latent correlation between the two modalities. Then a structure consolidator is proposed to integrate the structural information in KGs. The multimodal learner and structure consolidator are unified as a two-stage modality fusion strategy. Then we followed the principle of generative adversarial network to propose a relation embedding generator to learn the accurate representation for new relations based on their descriptions.
Experimental results on three datasets demonstrate the effectiveness of the proposed model in zero-shot relational learning in MMKGs, outperforming various baseline methods.
However, an limitation exists in our work. Despite some entities having multiple associated images, our model only exploits one image of them due to the restriction of an image-and-text input pair in the joint encoder. 
We will use multiple images to boost multimodal learning in future. 

\section{acknowledgement}
This work was supported by the National Science Foundation (award number 2333795) and the Healey Research Grant Program
of the University of Massachusetts Boston.

\bibliographystyle{IEEEtran}
\bibliography{main}

\end{document}